\documentclass[submission,copyright,creativecommons]{eptcs}

\usepackage[utf8]{inputenc}
\usepackage[T1]{fontenc}
\usepackage{textcomp}
\usepackage[english]{babel}
\usepackage{amsmath, amssymb}
\usepackage{tikz}
\usepackage{xfp, siunitx}
\usepackage{pgfplots}
\usetikzlibrary{arrows,positioning,matrix,backgrounds,calc,fit,decorations.pathreplacing,intersections,through}
\usepgfplotslibrary{statistics, fillbetween}
\pgfplotsset{compat=newest}
\pgfplotsset{height=4cm, width=7cm}
\pgfplotsset{
  layers/axis lines on top/.define layer set={
    axis background,
    axis grid,
    axis ticks,
    axis tick labels,
    pre main,
    main,
    axis lines,
    axis descriptions,
    axis foreground,
  }{/pgfplots/layers/standard},
}

\usepackage{import}
\usepackage{xifthen}
\usepackage{pdfpages}
\usepackage{transparent}

\newtheorem{prop}{Proposition}

\newcommand{\histogram}[3][]{
\begin{tikzpicture}
\begin{axis}[
    ymin=0,
    xmin=0, xmax=1,
    xtick={0,1/6,2/6,3/6,4/6,5/6,1},
    xticklabels={$0$,  $\frac{1}{6}$,  $\frac{2}{6}$,  $\frac{3}{6}$,  $\frac{4}{6}$,  $\frac{5}{6}$,  $1$},
    xlabel={Signalling fraction},
    minor y tick num = 4,
    grid,
    grid style=dashed,
    legend pos={outer north east},
    legend cell align={left},
    ]
\addplot+[ybar interval,mark=no,color=blue,fill=blue,fill opacity=0.3,ybar legend] plot coordinates {
#2 (0.16666667, 0)
};
\addplot+[ybar interval,mark=no,color=red,fill=red,fill opacity=0.3,ybar legend] plot coordinates {
#3 (1, 0)
};
\ifthenelse{\isempty{#1}}
{\legend{contextual, non-contextual}}
{}
\end{axis}
\end{tikzpicture}
}

\newcommand{\fourdp}[1]{\num[round-mode=places,round-precision=4]{#1}}
\newcommand{\model}[9][]{
\begin{tabular}{r|ccccc}
 #1 & $(\text{#2},\text{#2})$ & $(\text{#2}, \text{#3})$ & $(\text{#3}, \text{#2})$ & $(\text{#3}, \text{#3})$   \\ \hline
    $(\text{#4}, \text{#5})$ & $\fourdp{#7}$ & $0$ & $0$ & $\fourdp{\fpeval{1-#7}}$  \\
$(\text{#5}, \text{#6})$ & $\fourdp{#8}$ & $0$ & $0$ & $\fourdp{\fpeval{1-#8}}$  \\
$(\text{#6}, \text{#4})$ & $0$ & $\fourdp{#9}$ & $\fourdp{\fpeval{1-#9}}$ & $0$  
\end{tabular}
}
\begin{document}
\title{A Model of Anaphoric Ambiguities using Sheaf Theoretic Quantum-like Contextuality and BERT}
\def\titlerunning{Sheaves, BERT, and Anaphora }

\author{Kin Ian Lo\qquad \qquad Mehrnoosh Sadrzadeh
\institute{University College London \\London, UK}
\email{\{kin.lo.20,m.sadrzadeh\}@ucl.ac.uk}
\and
Shane Mansfield
\institute{Quandela \\ Paris, France}
\email{shane.mansfield@quandela.com}
}
\def\authorrunning{K. I. Lo,  M. Sadrzadeh \& S. Mansfield}

\maketitle              % typeset the header of the contribution
\begin{abstract}
Ambiguities of natural language do not preclude  us from using it and  context  helps in getting ideas across.  They, nonetheless, pose a key challenge to the development of competent machines to understand natural language and use it as humans do. 
Contextuality is an unparalleled phenomenon in quantum mechanics,  where  different mathematical formalisms have been put forwards to   understand and reason about it. 
In this paper, we construct a schema for anaphoric ambiguities that exhibits quantum-like contextuality. 
We use a recently developed criterion of sheaf-theoretic contextuality that is applicable to signalling models.
We then take advantage of the neural word embedding engine BERT to instantiate the schema to natural language examples and extract probability distributions for the instances. 
As a result, plenty of sheaf-contextual examples were discovered in the natural language corpora BERT utilises. Our hope is that these examples will pave the way for future research and for finding ways to extend  applications of quantum computing to natural language processing.

%\keywords{Coreference Ambiguities \and Anaphora \and Sheaf Theory  \and Signalling Fraction \and Quantum Contextuality \and BERT}
\end{abstract}

\section{Introduction}
Context plays a central role in determining   meanings of words, as words that often occur in similar contexts have similar meanings.  Conjured in the 1950's  by  Firth \cite{firth1957synopsis} and Harris \cite{harris1954distributional}, this hypothesis led to the field of \emph{Distributional} semantics. Harris noticed that words such as `eye doctor' and `optometrist'  occur in the same contexts, e.g. in the  neighbourhood of `eye' and `glasses'. Firth's infamous quote was that you know a word by the company it keeps.  The Distributional hypothesis has been formalised by vector semantics and implemented on large corpora of data. Originally, corpora of text were mined to build term-term co-occurrence matrices.   Nowadays, contextualised deep neural network architectures  such as {\bf BERT} are used to train the vector statistics.   Despite the daily successes of contextualised  embeddings in Natural Language Processing tasks, they do not have an explicit notion of  grammatical or discourse structure.  The statistics learnt by engines such as BERT do indeed take some structure into account when training  vector embeddings, but they certainly do not represent the grammatical or discourse structures of a piece of text in a vector in the same as they do for words. Compositional distributional semantics \cite{baroni2014frege,coecke2010mathematical} is a field of research introduced in an attempt to  address this challenge. A line of research of this field showed that by generalising the notion of vectors to tensors, one can embed both word and grammar. Recent research, has showed how quantum computing devices such as the IBMQ's quantum devices can be used to learn these tensors as quantum states \cite{Lorenzetal}. 

The links between natural language and quantum physics goes beyond the above.  Discovery of scenarios such as  EPR \cite{EPR} and Bell \cite{bell},  made quantum mechanics   the first science to formally deal with the notion of contextuality.   Scientists argued that quantum theory should be contextual in order to be sound and different mathematical formalisms were introduced to analyse this. Quantum-like contextuality  turned out to be, essentially,   the failure of having a global explanation to local observations on a system and the presence of   incompatible observables, in the sense  that a simultaneous global observation of all observables are not possible, except in trivial systems.  Over the last number of years it has been proved that it is this feature of quantum mechanics that is capable of lifting linear
computation to universal computation \cite{Anders2009, Raussendorf2013,
Mansfield2018} and that contextuality is necessary for magic state distillation
\cite{Howard2014}, a key component in fault-tolerant quantum computing schemes.
Roughly speaking, contextual systems hold additional computational power which is absent in non-contextual systems.   It is therefore a reasonable conjecture that quantum computers are better at dealing with contextual systems compared to classical computers. There exist a number of different frameworks for treating contextuality. The sheaf-theoretic framework of \cite{Abramsky2011} is amongst the ones that connects the statistical data collected from quantum experiments to the structures defined by quantum mechanics. One of these laws is the  \emph{no-signalling}  property, The sheaf-theoretic framework can only formalise contextual scenarios that are no-signalling. However, the examples we are aiming to study are highly unlikely to be non-signalling. We remedy this by using a recent extension of them to realistic experiments \cite{Emeriau2022}, where the authors derive a new inequality  to check the contextuality of systems of measurements with signalling data. In this setting,  some degree of signalling becomes possible.  We use this inequality to check the contextuality of our examples. 

Quantum-like contextuality has been observed in other fields, e.g. in behavioural sciences \cite{Dzhafarov2015} and natural language \cite{Wang2021a,Wang2021b}. In  \cite{Wang2021a,Wang2021b} Wang et al showed that pairs of ambiguous words can produce contextual systems that resemble the Bell/CHSH quantum measurement scenario. In this paper, we propose a novel linguistic construction that exhibits the contextuality of Coreference ambiguities and exemplify it to anaphoric relations. We use BERT to instantiate the construction and  extract probability distributions for  the instances. Checking the contextuality fraction for these instances showed that it is possible to discover examples of anaphoric ambiguity that exhibit quantum-like contextuality properties. In fact, we were able to find hundreds of examples after only working with a few pairs of nouns and their corresponding adjectives, verbs, and prepositional phrases.  We hope  finding contextual schema and instances in natural language data help us devise new quantum algorithms that can handle ambiguities better than classical computers do in natural language processing tasks. 

\section{Ambiguities in Natural Language}

One of the  ambiguities of natural languages comes from the fact that words  have different meanings.   For example,  `bat' has an animal meaning and a sport meaning,  `plant' can mean a living organism such as a tree or a shrub, or a manufacturing industrial unit, such as a power plant. \emph{Word Sense Disambiguation} is a long standing task and evaluation method in Natural Language Processing.  Here the goal is to identify which meaning  of a word is being used in a context.  Another major ambiguity in natural language comes from the \emph{Coreference Resolution} task:  the task of deciding which discourse entity is referring to which expression in a context. This is an important part of language engines such as dialogue systems or question answering. For instance, in an automatic MOT booking  system, the NLP engine should know which car the user is referring to when they say `I have a Toyota RAV4 and a Toyota Aqua, it is the hybrid one for which I need an MOT today.'.

Different  Coreference Resolution algorithms focus on different classes of referring expressions. Pronouns are in the class of definite referents and refer to entities that are identifiable from the context, because they have been mentioned before (or after). In the  discourse `Dawn called the AA. The car had broken down and she had no choice',  the pronoun `she' refers to the definite noun phrase `Dawn' and is an instance of the linguistic phenomena  \emph{anaphora}. Despite presence of linguistics properties in the anaphoric relations, such as gender and number agreement and grammatical role and verb preferences, these are  ambiguities. In `Dawn texted Wendy. Her car had broken down.',  or `Dawn phoned Wendy. She was upset and needed help.', it  is not clear whose car was broken or who was upset.   Pronominal coreference relations are many-to-many and the ambiguities arise from them taking complex forms.  A  pronoun can refer to multiple referents and multiple pronouns can refer to the same referent. In the  discourse `There is a man carrying a boy. He is tired and worn out. He is snoring.', the first \emph{He} can refer to both \emph{man} or \emph{boy}, but the second \emph{He} most certainly refers to \emph{boy}.

The different choices that give rise to ambiguities, be it in the choice of the meaning of a word in a Word Sense Disambiguation task, or the choice of the potential referent of an expression in a Coreference Resolution task, give rise to probability distributions. 
An ambiguous word can be treated as an observable which can have possible outcomes. In case of meaning ambiguities, these outcomes are possibilities over the semantic interpretations of the word.
A probability distribution over the outcomes can then be defined using the frequency of occurrences of the possible interpretations in a corpus, e.g. the entire English Wikipedia, or in plausibility judgements of human subjects.  A single observable is not sufficient to support contextuality, instead pairs of ambiguous words are needed. A pair of words is thought of as a pair of compatible observables measured simultaneously. 

The work of  \cite{Wang2021a,Wang2021b} focused on meaning ambiguities.  In this paper, we focus on coreference ambiguities and model contextual features of   ambiguities  arising from anaphoric  reference relations. An  identical approach can be taken if the relationship is cataphoric. We treat the pronouns as observables, of which the measurement outcomes are the possible referents of each pronoun. In what follows we describe the mathematical  setting we used, detail how to use it to model ambiguous anaphoric references,  explain how we found contextual examples, and present some of the contextual examples. 

\section{Sheaf Theoretic Framework}
In the sheaf-theoretic framework of contextuality \cite{Abramsky2011}, a measurement scenario is  a tuple $\langle \mathcal{X}, \mathcal{M}, \mathcal{O} \rangle$ with the data  $\cal X$, a set of observables,  $\cal M$, a measurement cover, and $\cal O$, a set of measurement outcomes. 
An observable in $\mathcal{X}$ is a quantity that can be measured to give one of the  outcomes in $ \mathcal{O} $. 
A subset of simultaneously measurable observables of $\mathcal{X}$ is called a measurement context (or simply called a context).  
The measurement cover $\mathcal{M}$ is a collection of contexts which covers $\mathcal{X}$, i.e. the union of all contexts in $\mathcal{M}$ is $ \mathcal{X}$.

For every measurement context, we can perform a number of repeated simultaneously measurements on the observables in the context. The gathered statistics can then be used to reconstruct an estimated joint probability distribution. 
Instead, one can also calculate the joint distribution exactly using an underlying theory of the concerned system, e.g. using Born's rule in quantum mechanics for a quantum system.

An \emph{empirical model} refers to a collection of such joint probability distribution for each context in the measurement cover $\mathcal{M}$.
By definition, a subset of observables in $\mathcal{X}$ that are not all included in a measurement context in $\mathcal{M}$ cannot be measured simultaneously. 
Therefore, a joint distribution over the said observables cannot be empirically estimated. 
The empirical model of a system fully encapsulates what is to be known from the system with empirical measurements. 

Contextuality comes from the failure of explaining an empirical model in a classically intuitive way: assuming that all measurements are just revealing deterministic pre-existing values, in other words, the measurement outcomes are already fixed when the system was prepared. 
Thus the randomness comes entirely from the system preparation. That means that there is a global joint distribution over all the observables in the scenario, which marginalises to every local joint distribution in the empirical model. 
Given an empirical model, if such a global distribution does not exist, then we call such empirical model contextual.
Note that such global distribution exists only in theory as there are observables in $ \mathcal{X}$ that cannot be measured simultaneously, unless in trivial scenarios. 

For readers familiar with sheaf theory, the said criterion for contextuality can be formalised using the language of sheaf. 
Consider the presheaf $ \mathcal{F}$ which assigns each subset $U \in \mathcal{P}(\mathcal{X})$ the set of all possible probability distributions on the observables in $U$. 
Each set inclusion $U \subseteq U'$, interpreted as an arrow in the category $\mathcal{P}(\mathcal{X})$, is mapped to the marginalisation of distributions on $U'$ to distributions on  $U$. 
For a measurement cover $ \mathcal{M}$, an empirical model is just a family of compatible distributions $ \{D_C\}_{C \in M} $.
The presheaf $\mathcal{F}$ is a sheaf if the gluing  property is satisfied:
\begin{quote}
    Fix a cover $ \mathcal{M}$ of $ \mathcal{X}$. For each family of compatible sections $\{D_C\}_{C \in \mathcal{M}} $, there is a unique global distribution compatible with every distributions in  $\{D_C\}_{C \in \mathcal{M}} $.
\end{quote}
Thus a contextual empirical model can only live on a measurement scenario for which the presheaf $\mathcal{F}$ is not a sheaf, i.e. not satisfying the gluing property.
To say that there is a contextual model that lives on a measurement scenario is to say that the presheaf $ \mathcal{F}$ is not a sheaf on the scenario.

As an example, the Bell/CHSH scenario involves two experimenters, Alice and Bob, who share between them a two-qubit quantum state. 
Alice is allowed to measure her part of the state with one of two incompatible observables, $ a_1$ and $ a_2$, which gives either $0$ or $1$ as the outcome.
Similarly Bob can choose to measure his part with observables $ b_1$ and $ b_2$.
Therefore, the Bell/CHSH measurement scenario is fully described with the following data: $\mathcal{X} = \{a_1, b_1, a_2, b_2\}$,  $\mathcal{M} = \{\{a_1,b_1\}, \{a_1, b_2\}, \{a_2,b_1\}, \{a_2, b_2\}    \}$, and $ \mathcal{O} = \{0, 1\} $.  Notice that $\{a_1, a_2\} $ and $\{b_1, b_2\} $ are not in $ \mathcal{M}$ as they cannot be measured simultaneously due to their quantum mechanical incompatibility.

So far we have specified what measurements are allowed and what outcomes are possible.
Suppose now Alice and Bob repeat the experiment many times and have gathered sufficient statistics to estimate the joint probability distribution for each context in $ \mathcal{M}$. Their results can be summarised in a table referred to as an \emph{empirical table}, see Figure \ref{fig:emptables}, where each row in the table represents a joint distribution on the context shown in the leftmost column. 
For instance, the bottom right entry in the table ($1 / 8$) is the probability of both Alice and Bob getting 1 as their measurement outcomes when Alice chooses to measure $ a_2$ and Bob chooses to measure $ b_2$.
Note that the empirical model of the system is entirely described by the empirical table.

\begin{figure}
\begin{center}
\begin{tabular}{r|ccccc}
 & $(0, 0)$ & $(0, 1)$ & $(1, 0)$ & $(1, 1)$   \\ \hline
$(a_1, b_1)$ & $1 / 2$ & $0$ & $0$ & $1 / 2$  \\
$(a_1, b_2)$ & $3 / 8$ & $1 / 8$ & $1 / 8$ & $3 / 8$  \\
$(a_2, b_1)$ & $3 / 8$ & $1 / 8$ & $1 / 8$ & $3 / 8$  \\
$(a_2, b_2)$ & $1 / 8$ & $3 / 8$ & $3 / 8$ & $1 / 8$  \\
\end{tabular}
\qquad
\begin{tabular}{r|ccccc}
 & $(0, 0)$ & $(0, 1)$ & $(1, 0)$ & $(1, 1)$   \\ \hline
$(a_1, b_1)$ & $1$ & $0$ & $0$ & $1$  \\
$(a_1, b_2)$ & $1$ & $1$ & $1$ & $1$  \\
$(a_2, b_1)$ & $1$ & $1$ & $1$ & $1$  \\
$(a_2, b_2)$ & $1$ & $1$ & $1$ & $1$  \\
\end{tabular}
\end{center}
\caption{Empirical tables of measurement scenarios:  Bell/CHSH (left), possibilistic Bell/CHSH  (right)}
\label{fig:emptables}
\end{figure}

One can show that, using elementary linear algebra, there exists no global distribution over $ \{a_1, a_2, b_1, b_2\} $ that marginalises to the 4 local distribution shown in the above empirical table.
Therefore, the empirical model considered here is indeed contextual.

Instead of probability, one can also consider possibility, i.e. whether an outcome is possible or not. 
If we use Boolean values to represent possibility, 0 for \emph{impossible} and 1 for \emph{possible}, the passage from probability to possibility is just a mapping of all zero probabilities to 0 and all non-zero probabilities to 1.
This (irreversible) mapping is called a \emph{possibilistic collapse} of the model.  For the empirical table of the possibilistic version of  Bell/CHSH see Figure \ref{fig:emptables}.  One can  visualise a possibilistic model with a bundle diagram, see Figure \ref{fig:bundles}:

\begin{figure}
\begin{center}
\begin{tikzpicture}[x=45pt,y=45pt,thick,label distance=-0.25em,baseline=(O.base), scale=0.7]
\coordinate (O) at (0,0);
\coordinate (T) at (0,1.5);
\coordinate (u) at (0,0.5);
\coordinate [inner sep=0em] (v0) at ($ ({-cos(1*pi/12 r)*1.2},{-sin(1*pi/12 r)*0.48}) $);
\coordinate [inner sep=0em] (v1) at ($ ({-cos(7*pi/12 r)*1.2},{-sin(7*pi/12 r)*0.48}) $);
\coordinate [inner sep=0em] (v2) at ($ ({-cos(13*pi/12 r)*1.2},{-sin(13*pi/12 r)*0.48}) $);
\coordinate [inner sep=0em] (v3) at ($ ({-cos(19*pi/12 r)*1.2},{-sin(19*pi/12 r)*0.48}) $);
\coordinate [inner sep=0em] (v0-0) at ($ (v0) + (T) $);
\coordinate [inner sep=0em] (v0-1) at ($ (v0-0) + (u) $);
\coordinate [inner sep=0em] (v1-0) at ($ (v1) + (T) $);
\coordinate [inner sep=0em] (v1-1) at ($ (v1-0) + (u) $);
\coordinate [inner sep=0em] (v2-0) at ($ (v2) + (T) $);
\coordinate [inner sep=0em] (v2-1) at ($ (v2-0) + (u) $);
\coordinate [inner sep=0em] (v3-0) at ($ (v3) + (T) $);
\coordinate [inner sep=0em] (v3-1) at ($ (v3-0) + (u) $);
\draw (v0) -- (v1) -- (v2) -- (v3) -- (v0);
\draw [dotted] (v0-1) -- (v0);
\draw [dotted] (v1-1) -- (v1);
\draw [dotted] (v2-1) -- (v2);
\draw [dotted] (v3-1) -- (v3);
\node [inner sep=0.1em] (v0') at (v0) {$\bullet$};
\node [anchor=east,inner sep=0em] at (v0'.west) {$a_1$};
\node [inner sep=0.1em,label={[label distance=-0.625em]330:{$b_1$}}] at (v1) {$\bullet$};
\node [inner sep=0.1em] (v2') at (v2) {$\bullet$};
\node [anchor=west,inner sep=0em] at (v2'.east) {$a_2$};
\node [inner sep=0.1em,label={[label distance=-0.5em]175:{$b_2$}}] at (v3) {$\bullet$};

%%% start - sections
\draw [line width=3.2pt,white] (v3-0) -- (v0-0);
\draw [line width=3.2pt,white] (v3-0) -- (v0-1);
\draw [line width=3.2pt,white] (v3-1) -- (v0-0);
\draw [line width=3.2pt,white] (v3-1) -- (v0-1);
\draw (v3-0) -- (v0-0);
\draw (v3-0) -- (v0-1);
\draw (v3-1) -- (v0-0);
\draw (v3-1) -- (v0-1);

\draw [line width=3.2pt,white] (v2-0) -- (v3-0);
\draw [line width=3.2pt,white] (v2-0) -- (v3-1);
\draw [line width=3.2pt,white] (v2-1) -- (v3-0);
\draw [line width=3.2pt,white] (v2-1) -- (v3-1);
\draw (v2-0) -- (v3-0);
\draw (v2-0) -- (v3-1);
\draw (v2-1) -- (v3-0);
\draw (v2-1) -- (v3-1);

\draw [line width=3.2pt,white] (v0-0) -- (v1-0);
%\draw [line width=3.2pt,white] (v0-0) -- (v1-1);
%\draw [line width=3.2pt,white] (v0-1) -- (v1-0);
\draw [line width=3.2pt,white] (v0-1) -- (v1-1);
\draw (v0-0) -- (v1-0);
%\draw (v0-0) -- (v1-1);
%\draw (v0-1) -- (v1-0);
\draw (v0-1) -- (v1-1);

\draw [line width=3.2pt,white] (v1-0) -- (v2-0);
\draw [line width=3.2pt,white] (v1-0) -- (v2-1);
\draw [line width=3.2pt,white] (v1-1) -- (v2-0);
\draw [line width=3.2pt,white] (v1-1) -- (v2-1);
\draw (v1-0) -- (v2-0);
\draw (v1-0) -- (v2-1);
\draw (v1-1) -- (v2-0);
\draw (v1-1) -- (v2-1);
%%% end - sections

\node [inner sep=0.1em,label=left:{$0$}] at (v0-0) {$\bullet$};
\node [inner sep=0.1em,label=left:{$1$}] at (v0-1) {$\bullet$};
\node [inner sep=0.1em,label={[label distance=-0.5em]330:{$0$}}] at (v1-0) {$\bullet$};
\node [inner sep=0.1em] at (v1-1) {$\bullet$};
\node [inner sep=0.1em,label=right:{$0$}] at (v2-0) {$\bullet$};
\node [inner sep=0.1em,label=right:{$1$}] at (v2-1) {$\bullet$};
\node [inner sep=0.1em] at (v3-0) {$\bullet$};
\node [inner sep=0.1em,label={[label distance=-0.5em]150:{$1$}}] at (v3-1) {$\bullet$};
\end{tikzpicture}
\qquad
\begin{tikzpicture}[x=45pt,y=45pt,thick,label distance=-0.25em,baseline=(O.base), scale=0.7]
\coordinate (O) at (0,0);
\coordinate (T) at (0,1.5);
\coordinate (u) at (0,0.5);
\coordinate [inner sep=0em] (v0) at ($ ({-cos(1*pi/12 r)*1.2},{-sin(1*pi/12 r)*0.48}) $);
\coordinate [inner sep=0em] (v1) at ($ ({-cos(7*pi/12 r)*1.2},{-sin(7*pi/12 r)*0.48}) $);
\coordinate [inner sep=0em] (v2) at ($ ({-cos(13*pi/12 r)*1.2},{-sin(13*pi/12 r)*0.48}) $);
\coordinate [inner sep=0em] (v3) at ($ ({-cos(19*pi/12 r)*1.2},{-sin(19*pi/12 r)*0.48}) $);
\coordinate [inner sep=0em] (v0-0) at ($ (v0) + (T) $);
\coordinate [inner sep=0em] (v0-1) at ($ (v0-0) + (u) $);
\coordinate [inner sep=0em] (v1-0) at ($ (v1) + (T) $);
\coordinate [inner sep=0em] (v1-1) at ($ (v1-0) + (u) $);
\coordinate [inner sep=0em] (v2-0) at ($ (v2) + (T) $);
\coordinate [inner sep=0em] (v2-1) at ($ (v2-0) + (u) $);
\coordinate [inner sep=0em] (v3-0) at ($ (v3) + (T) $);
\coordinate [inner sep=0em] (v3-1) at ($ (v3-0) + (u) $);
\draw (v0) -- (v1) -- (v2) -- (v3) -- (v0);
\draw [dotted] (v0-1) -- (v0);
\draw [dotted] (v1-1) -- (v1);
\draw [dotted] (v2-1) -- (v2);
\draw [dotted] (v3-1) -- (v3);
\node [inner sep=0.1em] (v0') at (v0) {$\bullet$};
\node [anchor=east,inner sep=0em] at (v0'.west) {$a_1$};
\node [inner sep=0.1em,label={[label distance=-0.625em]330:{$b_1$}}] at (v1) {$\bullet$};
\node [inner sep=0.1em] (v2') at (v2) {$\bullet$};
\node [anchor=west,inner sep=0em] at (v2'.east) {$a_2$};
\node [inner sep=0.1em,label={[label distance=-0.5em]175:{$b_2$}}] at (v3) {$\bullet$};

%%% start - sections
\draw [line width=3.2pt,white] (v3-0) -- (v0-0);
%\draw [line width=3.2pt,white] (v3-0) -- (v0-1);
%\draw [line width=3.2pt,white] (v3-1) -- (v0-0);
\draw [line width=3.2pt,white] (v3-1) -- (v0-1);
\draw (v3-0) -- (v0-0);
%\draw (v3-0) -- (v0-1);
%\draw (v3-1) -- (v0-0);
\draw (v3-1) -- (v0-1);

%\draw [line width=3.2pt,white] (v2-0) -- (v3-0);
\draw [line width=3.2pt,white] (v2-0) -- (v3-1);
\draw [line width=3.2pt,white] (v2-1) -- (v3-0);
%\draw [line width=3.2pt,white] (v2-1) -- (v3-1);
%\draw (v2-0) -- (v3-0);
\draw (v2-0) -- (v3-1);
\draw (v2-1) -- (v3-0);
%\draw (v2-1) -- (v3-1);

\draw [line width=3.2pt,white] (v0-0) -- (v1-0);
%\draw [line width=3.2pt,white] (v0-0) -- (v1-1);
%\draw [line width=3.2pt,white] (v0-1) -- (v1-0);
\draw [line width=3.2pt,white] (v0-1) -- (v1-1);
\draw (v0-0) -- (v1-0);
%\draw (v0-0) -- (v1-1);
%\draw (v0-1) -- (v1-0);
\draw (v0-1) -- (v1-1);

\draw [line width=3.2pt,white] (v1-0) -- (v2-0);
%\draw [line width=3.2pt,white] (v1-0) -- (v2-1);
%\draw [line width=3.2pt,white] (v1-1) -- (v2-0);
\draw [line width=3.2pt,white] (v1-1) -- (v2-1);
\draw (v1-0) -- (v2-0);
%\draw (v1-0) -- (v2-1);
%\draw (v1-1) -- (v2-0);
\draw (v1-1) -- (v2-1);
%%% end - sections

\node [inner sep=0.1em,label=left:{$0$}] at (v0-0) {$\bullet$};
\node [inner sep=0.1em,label=left:{$1$}] at (v0-1) {$\bullet$};
\node [inner sep=0.1em,label={[label distance=-0.5em]330:{$0$}}] at (v1-0) {$\bullet$};
\node [inner sep=0.1em] at (v1-1) {$\bullet$};
\node [inner sep=0.1em,label=right:{$0$}] at (v2-0) {$\bullet$};
\node [inner sep=0.1em,label=right:{$1$}] at (v2-1) {$\bullet$};
\node [inner sep=0.1em] at (v3-0) {$\bullet$};
\node [inner sep=0.1em,label={[label distance=-0.5em]150:{$1$}}] at (v3-1) {$\bullet$};
\end{tikzpicture}
\qquad
\begin{tikzpicture}[x=45pt,y=45pt,thick,label distance=-0.25em,baseline=(O.base), scale=0.7]
\coordinate (O) at (0,0);
\coordinate (T) at (0,1.5);
\coordinate (u) at (0,0.5);
\coordinate [inner sep=0em] (v0) at ($ ({-cos(-2*pi/12 r)*1.2},{-sin(-2*pi/12 r)*0.48}) $);
\coordinate [inner sep=0em] (v1) at ($ ({-cos(6*pi/12 r)*1.2},{-sin(6*pi/12 r)*0.48}) $);
\coordinate [inner sep=0em] (v2) at ($ ({-cos(14*pi/12 r)*1.2},{-sin(14*pi/12 r)*0.48}) $);
\coordinate [inner sep=0em] (v0-0) at ($ (v0) + (T) $);
\coordinate [inner sep=0em] (v0-1) at ($ (v0-0) + (u) $);
\coordinate [inner sep=0em] (v1-0) at ($ (v1) + (T) $);
\coordinate [inner sep=0em] (v1-1) at ($ (v1-0) + (u) $);
\coordinate [inner sep=0em] (v2-0) at ($ (v2) + (T) $);
\coordinate [inner sep=0em] (v2-1) at ($ (v2-0) + (u) $);
\draw (v0) -- (v1) -- (v2) -- (v0);
\draw [dotted] (v0-1) -- (v0);
\draw [dotted] (v1-1) -- (v1);
\draw [dotted] (v2-1) -- (v2);

\node [inner sep=0.1em] (v0') at (v0) {$\bullet$};
\node [anchor=east,inner sep=0em] at (v0'.west) {$x_1$};
\node [inner sep=0.1em,label={[label distance=-0.625em]330:{$x_2$}}] at (v1) {$\bullet$};
\node [inner sep=0.1em,label={[label distance=-0.0em]0:{}}] (v2') at (v2) {$\bullet$};
\node [anchor=west,inner sep=0em] at (v2'.east) {$x_3$};

%%% start - sections
\draw [line width=3.2pt,white] (v0-1) -- (v2-1);
%\draw [line width=3.2pt,white] (v0-1) -- (v2-0);
%\draw [line width=3.2pt,white] (v0-0) -- (v2-1);
\draw [line width=3.2pt,white] (v0-0) -- (v2-0);
\draw (v0-1) -- (v2-1);
%\draw (v0-1) -- (v2-0);
%\draw (v0-0) -- (v2-1);
\draw (v0-0) -- (v2-0);

\draw [line width=3.2pt,white] (v2-1) -- (v1-1);
%\draw [line width=3.2pt,white] (v2-1) -- (v1-0);
%\draw [line width=3.2pt,white] (v2-0) -- (v1-1);
\draw [line width=3.2pt,white] (v2-0) -- (v1-0);
\draw (v2-1) -- (v1-1);
%\draw (v2-1) -- (v1-0);
%\draw (v2-0) -- (v1-1);
\draw (v2-0) -- (v1-0);

%\draw [line width=3.2pt,white] (v1-1) -- (v0-1);
\draw [line width=3.2pt,white] (v1-1) -- (v0-0);
\draw [line width=3.2pt,white] (v1-0) -- (v0-1);
%\draw [line width=3.2pt,white] (v1-0) -- (v0-0);
%\draw (v1-1) -- (v0-1);
\draw (v1-1) -- (v0-0);
\draw (v1-0) -- (v0-1);
%\draw (v1-0) -- (v0-0);
%%% end - sections

\node [inner sep=0.1em,label=left:{$0$}] at (v0-0) {$\bullet$};
\node [inner sep=0.1em,label=left:{$1$}] at (v0-1) {$\bullet$};
\node [inner sep=0.1em,label={[label distance=-0.5em]330:{$0$}}] at (v1-0) {$\bullet$};
\node [inner sep=0.1em] at (v1-1) {$\bullet$};
\node [inner sep=0.1em,label=right:{$0$}] at (v2-0) {$\bullet$};
\node [inner sep=0.1em,label=right:{$1$}] at (v2-1) {$\bullet$};
\end{tikzpicture}
\end{center}
\caption{Bundle diagrams of possibilistic CHSH (left), PR box (middle),  PR prism (right)}
\label{fig:bundles}
\end{figure}
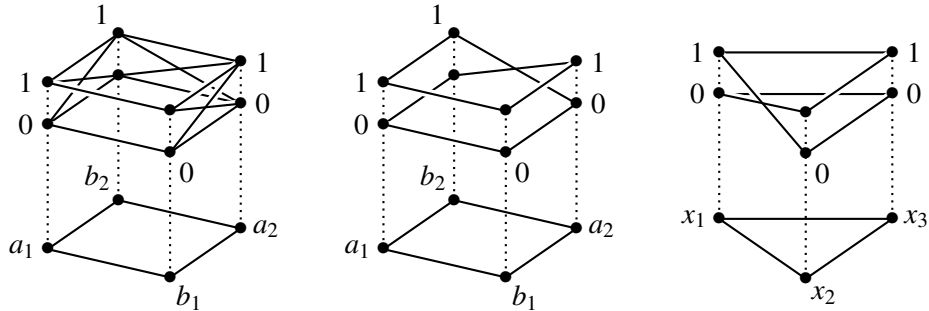
The base (i.e. the bottom part) of the bundle diagram represents the measurement cover $\mathcal{M}$, where each vertex represents an observable in $\mathcal{X}$.
An edge is drawn between two observables if they can be simultaneously measured, i.e. in the same measurement context. 
What sits on top of the base represents the possible outcomes. For instance, the presence of the edge connecting the 0 vertex on top of $ a_1$ and the 0 vertex on the observable $ b_1$ means that it is possible to get the joint outcome $(0, 0)$ when the context  $(a_1, b_1) $ is measured.

A system is logically contextual if the inexistence of a global distribution can already be deduced by looking at the supports of the context-wise distributions -- or
equivalently if the Boolean distributions obtained by the \emph{possibilistic
collapse} of the model \cite{Abramsky2011} is contextual. Such systems are said to be \emph{possibilistically contextual}.  
Logical contextuality manifests on a bundle diagram as the failure of extending at least one of the edges to a loop that wraps around the base once.  For the possibilistic empirical model of a PR box, see Figure \ref{fig:bundles}. 
Note that none of the edges is extendable to a loop that wraps around the base once. Given the possibilistic collapse of an empirical model, if none of the edges can be extendable to a loop that wraps around the base once, we say that the model is \emph{strongly contextual}.\footnote{Strictly speaking, this definition of strong contextuality only applies to cyclic scenarios where the base of the bundle diagram forms a loop. Nonetheless, cyclic scenarios are the only scenarios considered in this paper.}

\begin{prop}
The \emph{minimal measurement scenario} that admits contextuality has the data up to relabelling: 
   ${ \cal X} = \{x_1, x_2, x_3\}$, 
    ${\cal M} = \{\{x_1, x_2\}, \{x_2, x_3\},  \{x_1, x_3\}\}$, and 
    ${\cal O} = \{0,1\}$.
\end{prop}

\noindent
The proof of the above is routine, and so is that of the following: 

\begin{prop} The only strongly contextual system, up to relabelling, for the minimal measurement scenario is where perfect correlation is observed on two of the contexts and perfect anti-correlation is observed on the other one. 
\end{prop}

We call this scenario the \emph{PR prism} as an analogy to the PR boxes. See Figure \ref{fig:bundles} for its bundle diagram. The pairs of parallel edges over contexts $\{x_2, x_3\} $ and $\{x_3, x_1\} $ correspond to perfect correlation and the pair of crossed edges over context $ \{x_1, x_2\} $ corresponds to the perfect anti-correlation.

\section{Contextual and Signalling Fractions}

The contextual fraction (CF) \cite{Abramsky2017} measures the degree of contextuality of a given non-signalling model.
Given an empirical model $e$, the CF of $e$ is defined as the minimum $\lambda$ such that the following convex decomposition of $e$ works\footnote{Here, we represent the empirical models as empirical tables. Addition and scalar multiplication are then interpreted as standard matrix operations, where the empirical tables are treated as matrices.}:
\begin{equation}
    \label{eq:convexcf}
    e = (1-\lambda) e^{NC} + \lambda e^{C},
\end{equation}
where $e^{NC}$ is a non-contextual (and non-signalling) empirical model and  $e^{C}$ is a model allowed to be contextual. 
For non-signalling models, the criterion of contextuality is just
\begin{equation}
    \label{eq:cfnosig}
   \text{CF} > 0.
\end{equation}
As $e^{NC}$ is not allowed to be signalling, the CF of a signalling model must be greater than zero.
Thus, interpreting CF as a measure of contextuality for signalling models
would lead to erroneous conclusions. 
However, most models, including the ones considered in this paper, are signalling.

One can try to define a signalling fraction (SF), in the same way CF is defined, to quantify the degree of signalling. Given a model $e$, the SF of $e$ is defined as the minimum $\mu$ such that the following convex decomposition of  $e$ works:
\begin{equation}
    \label{eq:convexsf}
    e = (1-\mu) e^{NS} + \mu e^{S},
\end{equation}
where $e^{NS}$ is a non-signalling empirical model and  $e^{S}$ is a model allowed to be signalling.

In \cite{Emeriau2022}, the signalling fraction (SF) was used to quantify the amount of fictitious contextuality contributing to the contextual fraction due to signalling in a signalling model.
The authors derived a criterion of contextuality for signalling models that reads 
\begin{equation}
\label{eq:cf}
\text{CF} > 2 |\cal M| \, \text{SF}
,\end{equation}
where $|\cal M|$ denotes the number of measurement contexts. Notice how criterion (\ref{eq:cf}) reduces to the generalised criterion (\ref{eq:cfnosig}) when $\text{SF} = 0$.

In the general case, one would need to solve a linear program to calculate the CF or SF of a model. 
The calculation is much simpler with models that share the same support as the PR prism. 
We call these model \emph{PR-like}. Such models can always be written as the following empirical table upon relabelling:
\begin{center}
\begin{tabular}{r|ccccc}
 & $(0, 0)$ & $(0, 1)$ & $(1, 0)$ & $(1, 1)$   \\ \hline
$(x_1, x_2)$ & $(1+\epsilon_1) / 2$ & $0$ & $0$ & $(1-\epsilon_1) / 2$  \\
$(x_2, x_3)$ & $(1+\epsilon_2) / 2$ & $0$ & $0$ & $(1-\epsilon_2) / 2$  \\
$(x_3, x_1)$ & $0$ & $(1+\epsilon_3) / 2$ & $(1-\epsilon_3) / 2$ & $0$  
\end{tabular}
\end{center}
where $-1 \leq \epsilon_1, \epsilon_2, \epsilon_3 \leq 1$. 
Recall that the model $e^{NC}$ in the convex decomposition (\ref{eq:convexcf}) is noncontextual and non-signalling. 
For a PR-like model to be non-signalling, one can check that a PR-like model is non-signalling if and only if it is a PR box, i.e. $\epsilon_1 = \epsilon_2 = \epsilon_3 = 0$.
However, the PR box is known to be (strongly) contextual. Thus, there does not exist a model $e^{NC}$ that is noncontextual and non-signalling for a PR-like model. Therefore, the SF of PR-like model is always $1$.

The calculation of SF for PR-like models is also simple. 
As $e^{NS}$ in the convex decomposition (\ref{eq:convexsf}) can be contextual but not signalling, $e^{NS}$ must be the PR box, the one with $\epsilon_1 = \epsilon_2=\epsilon_3=0$. 
As we cannot have negative probabilities in $e^{S}$ in the decomposition, the coefficient $(1-\mu)$ can at most be double the smallest non-zero value in the table, that is, $\min(1 \pm \epsilon_i)$. Thus we have 
\begin{equation*}
    \label{eq:sfprlike}
    SF = 1 - \min_{i=1,2,3}(1 \pm \epsilon_i) = \max_{i=1,2,3}|\epsilon_i| 
\end{equation*}
for PR-like models. 
We will use this result to calculate the SF of the PR-like models we constructed in the following section.

\begin{figure}[t]
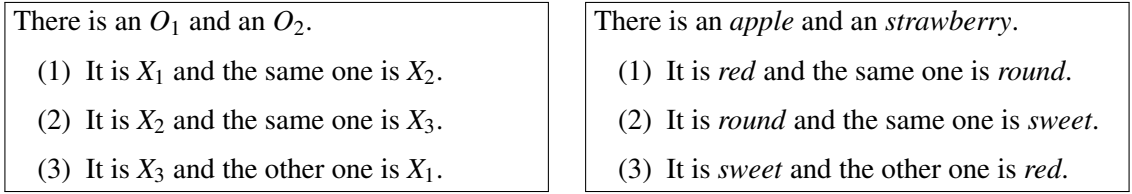

\begin{center}
\fbox{\begin{minipage}{7cm}{
 There is an $O_1$ and an $O_2$. 
 \begin{enumerate}
     \item[(1)] It is $ X_1$ and the same one is $ X_2$. 
     \item[(2)] It  is $ X_2$ and  the same one is $ X_3$. 
     \item[(3)] It is  $X_3$ and the other one is $ X_1$.
 \end{enumerate}}\end{minipage}}
 \quad
 \fbox{\begin{minipage}{7cm}{
     There is an \emph{apple} and an \emph{strawberry}. 
 \begin{enumerate}
     \item[(1)] It is \emph{red} and the same one is \emph{round}. 
     \item[(2)]  It  is \emph{round} and the same one is  \emph{sweet}. 
     \item[(3)] It is  \emph{sweet} and the other one is \emph{red}.
 \end{enumerate}}\end{minipage}}
 \end{center}
 \caption{The PR prism schema and its adjective modifier instance.}
 \label{fig:schemadj}
 \end{figure}

\section{Possibilistic  Examples}
\label{sec:pos}

The construction used in a previous work on meaning ambiguities \cite{Wang2021a} was inspired by the Bell/CHSH scenario in quantum physics. 
However, the Bell/CHSH scenario is not minimal so we considered the minimal scenario with only 3 observables instead of 4.
In our anaphoric setting, the set of possible interpretations is dependent on the ambiguous anaphora, instead of a fixed set of interpretations in the case of meaning ambiguities. 
This poses a difficulty in obtaining  probabilities through a corpus. So we first focus on possibility instead of probability. It is much easier to determine if it makes sense for a word to be the referent of an anaphora than to determine its likelihood. 
We constructed a schema (Figure \ref{fig:schemadj}) that is modelled by the PR prism on the possibilistic level.

In the schema, $O_1$ and $O_2$ are two noun phrases as the candidate referents;  $X_1, X_2, X_3$ are three modifiers commonly used to act on  $ O_1, O_2$. The $X_i$'s  are the observables of the scenario.\footnote{The ambiguous anaphoric words in the schema are \emph{it} and \emph{one}. We acknowledge that it is controversial to treat the modifiers $X_i$, instead of the ambiguous words, as observables. The construction of a more natural sounding schema is left for future research.}   
Statement (1) and (2) above ensure that the modifiers $X_i$ refer to the same referent, thus resulting in perfect correlation (parallel edges on the bundle diagram).
Statement (3) ensures that the modifiers refer to different referents, thus resulting in perfect anti-correlation (crossing edges on the bundle diagram). The schema is constructed such that it is minimal and can immediately be modelled by the PR prism to ensure strong contextuality.

\begin{figure}[t]
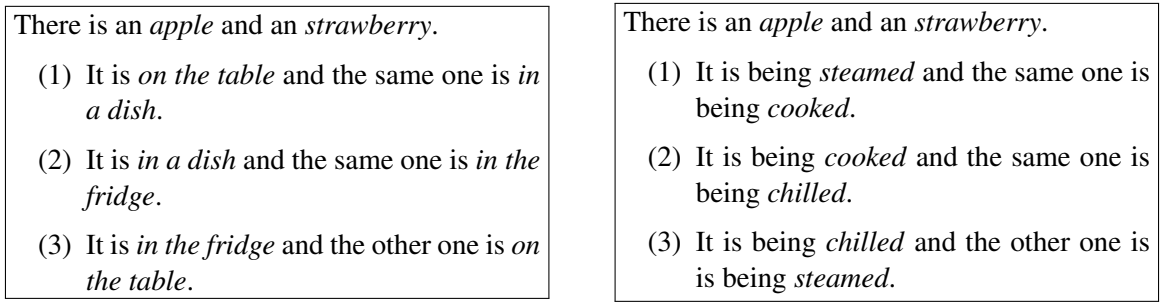

\centering
\fbox{\begin{minipage}{7cm}
    There is an \emph{apple} and an \emph{strawberry}. 
 \begin{enumerate}
     \item[(1)] It is \emph{on the table} and the same one is \emph{in a dish}. 
     \item[(2)]  It  is \emph{in a dish} and the same one is \emph{in the fridge}. 
     \item[(3)] It is  \emph{in the fridge} and the other one is \emph{on the table}.
 \end{enumerate}\end{minipage}}
 \qquad
 \fbox{\begin{minipage}{7cm}
    There is an \emph{apple} and an \emph{strawberry}. 
 \begin{enumerate}
     \item[(1)] It is being \emph{steamed} and the same one is being \emph{cooked}. 
     \item[(2)]  It  is being \emph{cooked} and the same one is being \emph{chilled}. 
     \item[(3)] It is being \emph{chilled} and the other one is is being \emph{steamed}.
 \end{enumerate}
\end{minipage}}
\caption{Examples of the PR prism schema with verbs (left) and preposition modifiers (right)}
\label{fig:Prism}
\end{figure}

\noindent For other examples using the same pair of nouns but with instead verbs or prepositional modifiers, see Figure \ref{fig:Prism}. Other types of modifier are dealt with similarly.

\section{Probabilistic  Examples}
\label{sec:prob}
We considered possibilistic models in the previous section.  In this section, we propose a method for defining probability distributions for schemas such as the one considered in the previous sections. 

We form a probabilistic model through a contextualised language model such as BERT \cite{Devlin2019}, which predicts a masked word (i.e. a blank space) in a sentence.
Intuitively speaking, BERT uses the sentence as the context to generate a contextualised word embedding for the masked word, which is then measured for similarity against every word in the vocabulary.

For example, given a sentence:  {\small  \texttt{The goal of life is [MASK].}}, BERT predicts the most likely word in the place of \texttt{[MASK]}. Moreover, BERT  assigns a probability score to every word in the vocabulary. 

%One can restrict the range  to the top $n$ candidate words predicted by BERT for the \texttt{MASK}. In this case, a probability distributions over the $n$ candidates is obtained by normalising their corresponding probability scores.  The top 5 candidate words predicted by BERT for the \texttt{[MASK]} in the above sentence and their probability scores and normalised probability distribution are shown below, respectively. 
The top 5 candidate words predicted by BERT and their probability scores are shown below.

\begin{center}
\begin{tabular}{r|cccccc}
\emph{probability scores}   & \texttt{life} & \texttt{survival} & \texttt{love} & \texttt{freedom} & \texttt{simplicity} & $\cdots$ \\
   \hline
    Probability & 0.1093 & 0.0394 & 0.0329 & 0.0300 & 0.0249 & $\cdots$
\end{tabular}   
\end{center}

%\begin{center}
%\begin{tabular}{r|cccccc}
%\emph{probability distribution}   & \texttt{life} & \texttt{survival} & \texttt{love} & \texttt{freedom} & \texttt{simplicity}  \\
   %\hline
    %Probability & 0.4621 & 0.1665 & 0.1391 & 0.1268 & 0.1052 
%\end{tabular}   
%\end{center}

We choose to use BERT because it has been providing improved  baselines for many   NLP tasks. 
In the following, we will demonstrate how we used BERT to define a probabilistic model for every schema considered in the previous section.

Consider the \emph{apple}-\emph{strawberry}  example of Section \ref{sec:pos}. To measure a context, we replace the pronoun \emph{It} in the sentence with \texttt{The [MASK]}. 
In practice, we feed the following 3 sentences separately  to BERT:

\begin{minipage}{17cm}{
         {\small  \texttt{There is an apple and an strawberry. The [MASK] is red and the same one is round.}}\\
         {\small  \texttt{There is an apple and an strawberry. The [MASK] is round and the same one is sweet.}}\\
         {\small  \texttt{There is an apple and an strawberry. The [MASK] is sweet and the other one is red.}}
         }
\end{minipage}

\noindent BERT will then produce, probabilities $P_i\left( \texttt{apple} \right) $ and $P_i\left( \texttt{strawberry} \right) $ for the i-th sentence shown above. As BERT gives a probability score to every word in the vocabulary which sum to one, it is almost impossible that $P_i\left( \texttt{apple} \right) +P_i\left( \texttt{strawberry} \right) = 1$. 
We therefore normalise them by the following map\footnote{The normalisation here is equivalent to limiting the vocabulary to just \texttt{apple} and \texttt{strawberry} when BERT computes the probability scores.}:
\begin{align*}
    P_i\left( \texttt{apple} \right) \mapsto {P_i\left( \texttt{apple} \right)}/{(P_i\left( \texttt{apple} \right) + P_i\left( \texttt{strawberry} \right))} \\
    P_i\left( \texttt{strawberry} \right) \mapsto {P_i\left( \texttt{strawberry} \right)}/{(P_i\left( \texttt{apple} \right) + P_i\left( \texttt{strawberry} \right))}
\end{align*}
We will then use the normalised probabilities to construct a PR-like model with empirical table:
\begin{center}
\begin{tabular}{r|ccccc}
 & $(\text{apple},\text{apple})$ & $(\text{apple}, \text{strawberry})$ & $(\text{strawberry}, \text{apple})$ & $(\text{strawberry}, \text{strawberry})$   \\ \hline
$(\text{red}, \text{round})$ & $P_1\left( \texttt{apple} \right) $ & $0$ & $0$ & $P_1\left( \texttt{strawberry} \right) $  \\
$(\text{round}, \text{sweet})$ & $P_2\left( \texttt{apple} \right) $ & $0$ & $0$ & $P_2\left( \texttt{strawberry} \right) $  \\
$(\text{sweet}, \text{red})$ & $0$ & $P_3\left( \texttt{apple} \right) $ & $P_3\left( \texttt{strawberry} \right) $ & $0$  
\end{tabular}
\end{center}
It should be obvious how this procedure can be used on other examples of the schemas we considered in the last section.
Notice that such an empirical model is non-signalling only if $P_i\left( \texttt{apple} \right) = P_i\left( \texttt{strawberry} \right) = 0.5$ for all $i$.
It is therefore very unlikely that the model is non-signalling. To determine whether a signalling model is contextual, we use the inequality criterion of Equation (\ref{eq:cf}). 
Recall that the CF of a PR-like model is always 1.
Also, all the examples we considered in this paper have 3 contexts, i.e. $|\mathcal{M}| = 3$.
Thus, to tell if such a model is contextual, we just need to check if  $\text{SF} < \frac{1}{6}$. 

As the criterion is actually quite strict, it is unlikely for any model constructed in this way to be contextual. 
We therefore need to create plenty of examples and to be strategic in the way we construct them.
Equation (\ref{eq:sfprlike}) for PR-like models indicates that we need to make the probabilities as balanced as possible to make SF small.
For that, we first fix two semantically similar nouns or noun phrases. Then, we ask BERT to associate  them with frequently used modifying adjectives, verbs and prepositional phrases. As a result,  we can ensure that the probabilities for the masked word given by BERT will be relatively balanced and thus minimising signalling in the model. The examples that produce contextual empirical models are presented in the proceedings subsections. 

 \begin{figure}[t]
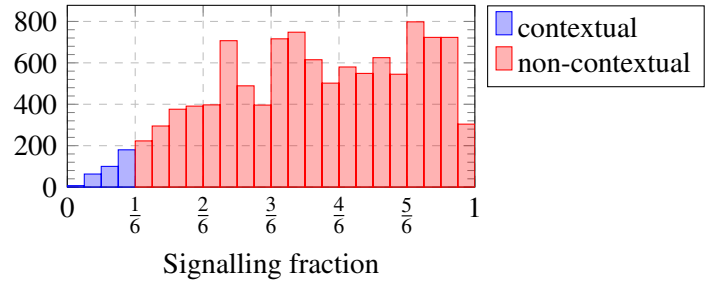

 \centering

\histogram{(0.0000, 7) (0.0417, 63) (0.0833, 100) (0.1250, 180) }{(0.1667, 223) (0.2083, 295) (0.2500, 376) (0.2917, 391) (0.3333, 397) (0.3750, 707) (0.4167, 489) (0.4583, 396) (0.5000, 716) (0.5417, 748) (0.5833, 615) (0.6250, 502) (0.6667, 580) (0.7083, 549) (0.7500, 625) (0.7917, 545) (0.8333, 798) (0.8750, 723) (0.9167, 723) (0.9583, 304) }
 \caption{The distribution of signalling fractions of the models constructed with adjective modifiers.}
 \label{fig:nosigfracadj}
\end{figure}

\subsection{Adjective Modifiers}
 
We considered 11 pairs of similar noun phrases with between 3 and 18 candidate adjectives respectively. 
A model is constructed by picking a triple of adjective modifiers as the observables from the list of adjectives shown in the Appendix. 
This data  generated 11,052 empirical models, of which 350 were contextual.  
Out of the 11 noun pairs considered, (cat, dog), (girl, boy) and (man, woman) produced models that are contextual.
See below for the empirical tables of 2 examples of the contextual models we found.
\begin{center}
\model[(1)]{cat}{dog}{good}{young}{small}{0.49405530095100403}{0.45355847477912903}{0.5718123912811279}
\end{center}
\begin{center}
    \model[(2)]{girl}{boy}{young}{small}{little}{0.5711000561714172}{0.5654845833778381}{0.52799391746521}
\end{center}

Figure \ref{fig:nosigfracadj} is a histogram of the distribution of signalling fractions of the models constructed using the adjective modifiers considered. 
One can see that the majority of the model constructed are non-contextual and that the distribution skews towards greater SF.

\subsection{Verb Phrases}
\begin{figure}[t]
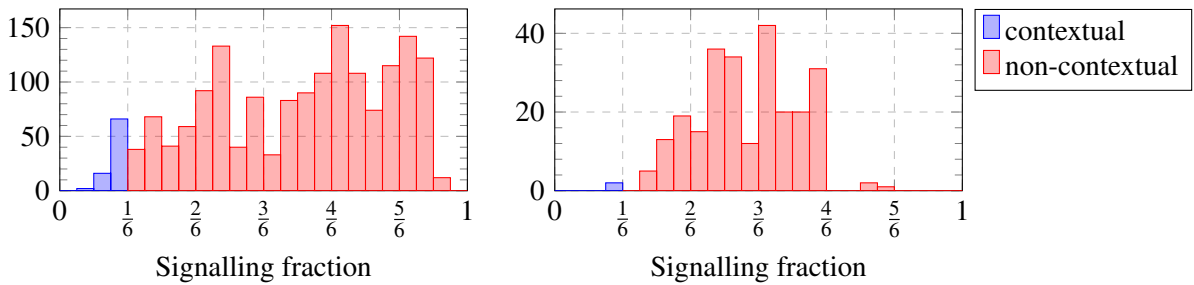

     \centering
     \histogram[no legend]{(0.0000, 0) (0.0417, 2) (0.0833, 16) (0.1250, 66) }{(0.1667, 38) (0.2083, 68) (0.2500, 41) (0.2917, 59) (0.3333, 92) (0.3750, 133) (0.4167, 40) (0.4583, 86) (0.5000, 33) (0.5417, 83) (0.5833, 90) (0.6250, 108) (0.6667, 152) (0.7083, 108) (0.7500, 74) (0.7917, 115) (0.8333, 142) (0.8750, 122) (0.9167, 12) (0.9583, 0) }
\histogram{(0.0000, 0) (0.0417, 0) (0.0833, 0) (0.1250, 2) }{(0.1667, 0) (0.2083, 5) (0.2500, 13) (0.2917, 19) (0.3333, 15) (0.3750, 36) (0.4167, 34) (0.4583, 12) (0.5000, 42) (0.5417, 20) (0.5833, 20) (0.6250, 31) (0.6667, 0) (0.7083, 0) (0.7500, 2) (0.7917, 1) (0.8333, 0) (0.8750, 0) (0.9167, 0) (0.9583, 0) }
\caption{The distributions of signalling fractions for verbs (left) and prepositions (right).}
 \label{fig:nosigfracverbprep}
\end{figure}

We considered 2 pairs of similar noun phrases with 8 and 9 candidate verbs respectively, see the table in Appendix for details. This data generated 1,680 empirical models, of which 84 were contextual. For instance, the empirical table of the (apple, strawberry) - (sold, eaten, chilled) contextual model is presented below.
The histogram of signalling fractions of the models constructed here is shown in the left panel of Figure \ref{fig:nosigfracverbprep}.
\begin{center}
\model{strawberry}{apple}{sold}{eaten}{chilled}{0.45867520570755005}{0.5621004104614258}{0.4415716826915741}
\end{center}

  \subsection{Prepositional Phrases}

We considered 2 pairs of noun phrases with 3 and 6 prepositional phrases respectively, see the Appendix. This data generated 252 empirical models, and we found two contextual models for each noun pair. For an example empirical see below; see the right panel of Figure \ref{fig:nosigfracverbprep} for the distribution of signalling fractions. Here \emph{strawberry} is abbreviated as \emph{strawb.} in the interest of space.
 \begin{center}
\model{apple}{strawb.}{on the table}{in the fridge}{in a dish}{0.5590721368789673}{0.563956618309021}{0.4777790307998657}
\end{center}

%An interesting observation was that any model that is sheaf theoretic contextual is also contextual in the Contextuality-by-Default (CbD) framework \cite{Dzhafarov2016}. As a result, it was easier to find CbD contextual examples and with our above data, we found \begin{color}{magenta} this many \end{color} of CbD contextual examnples. 

\section{Conclusions and Future Work}

Coreference resolution is, amongst other Natural Language Processing tasks, facing the challenge of  ambiguities.  Instances of this task require extra resources such as context and world knowledge.  
In this paper, we focused on anaphoric coreference relations and the role of context. 
We showed how realistic contextual sheaf theoretic models of ambiguous data arising from quantum-inspired scenarios \cite{Emeriau2022}  can be used to model these examples. 
We developed a schema that produces possibilistic contextual models analogous to the PR Box. 
We mined probabilities for the instances of this schema using the BERT neural language model.
Our computations showed that it is possible to find possibilistic as well as probabilistic contextual examples in natural language data, with only a handful of noun phrases and their modifiers. 
Future works include applying a similar methodology to coreference relations such as indefinite and definite noun phrases, quantifier scope, and situations requiring world knowledge, e.g. the Winograd Schema Challenge \cite{Levesque2012}.

\bibliographystyle{eptcs}
\bibliography{references.bib}

\begin{thebibliography}{10}
\providecommand{\bibitemdeclare}[2]{}
\providecommand{\surnamestart}{}
\providecommand{\surnameend}{}
\providecommand{\urlprefix}{Available at }
\providecommand{\url}[1]{\texttt{#1}}
\providecommand{\href}[2]{\texttt{#2}}
\providecommand{\urlalt}[2]{\href{#1}{#2}}
\providecommand{\doi}[1]{doi:\urlalt{https://doi.org/#1}{#1}}
\providecommand{\eprint}[1]{arXiv:\urlalt{https://arxiv.org/abs/#1}{#1}}
\providecommand{\bibinfo}[2]{#2}

\bibitemdeclare{article}{Abramsky2017}
\bibitem{Abramsky2017}
\bibinfo{author}{Samson \surnamestart Abramsky\surnameend},
  \bibinfo{author}{Rui~Soares \surnamestart Barbosa\surnameend} \&
  \bibinfo{author}{Shane \surnamestart Mansfield\surnameend}
  (\bibinfo{year}{2017}): \emph{\bibinfo{title}{Contextual Fraction as a
  Measure of Contextuality}}.
\newblock {\slshape \bibinfo{journal}{Physical Review Letter}}
  \bibinfo{volume}{119}, p. \bibinfo{pages}{050504},
  \doi{10.1103/PhysRevLett.119.050504}.

\bibitemdeclare{article}{Abramsky2011}
\bibitem{Abramsky2011}
\bibinfo{author}{Samson \surnamestart Abramsky\surnameend} \&
  \bibinfo{author}{Adam \surnamestart Brandenburger\surnameend}
  (\bibinfo{year}{2011}): \emph{\bibinfo{title}{The sheaf-theoretic structure
  of non-locality and contextuality}}.
\newblock {\slshape \bibinfo{journal}{New Journal of Physics}}
  \bibinfo{volume}{13}(\bibinfo{number}{11}), p. \bibinfo{pages}{113036},
  \doi{10.1088/1367-2630/13/11/113036}.

\bibitemdeclare{article}{Anders2009}
\bibitem{Anders2009}
\bibinfo{author}{Janet \surnamestart Anders\surnameend} \&
  \bibinfo{author}{Dan~E. \surnamestart Browne\surnameend}
  (\bibinfo{year}{2009}): \emph{\bibinfo{title}{Computational Power of
  Correlations}}.
\newblock {\slshape \bibinfo{journal}{Physical Review Letter}}
  \bibinfo{volume}{102}, p. \bibinfo{pages}{050502},
  \doi{10.1103/PhysRevLett.102.050502}.

\bibitemdeclare{inproceedings}{baroni2014frege}
\bibitem{baroni2014frege}
\bibinfo{author}{Marco \surnamestart Baroni\surnameend},
  \bibinfo{author}{Raffaella \surnamestart Bernardi\surnameend} \&
  \bibinfo{author}{Roberto \surnamestart Zamparelli\surnameend}
  (\bibinfo{year}{2014}): \emph{\bibinfo{title}{Frege in Space: A Program for
  Compositional Distributional Semantics}}.
\newblock \bibinfo{volume}{9}, p. \bibinfo{pages}{241–346},
  \doi{10.33011/lilt.v9i.1321}.

\bibitemdeclare{article}{bell}
\bibitem{bell}
\bibinfo{author}{John~S. \surnamestart Bell\surnameend} (\bibinfo{year}{1964}):
  \emph{\bibinfo{title}{{On the Einstein Podolsky Rosen paradox}}}.
\newblock {\slshape \bibinfo{journal}{Physics Physique Fizika}}
  \bibinfo{volume}{1}, pp. \bibinfo{pages}{195--200},
  \doi{10.1103/PhysicsPhysiqueFizika.1.195}.

\bibitemdeclare{article}{coecke2010mathematical}
\bibitem{coecke2010mathematical}
\bibinfo{author}{Bob \surnamestart Coecke\surnameend},
  \bibinfo{author}{Mehrnoosh \surnamestart Sadrzadeh\surnameend} \&
  \bibinfo{author}{Stephen \surnamestart Clark\surnameend}
  (\bibinfo{year}{2010}): \emph{\bibinfo{title}{Mathematical Foundations for a
  Compositional Distributional Model of Meaning}}.
\newblock \doi{10.48550/arXiv.1003.4394}.

\bibitemdeclare{inproceedings}{Devlin2019}
\bibitem{Devlin2019}
\bibinfo{author}{Jacob \surnamestart Devlin\surnameend},
  \bibinfo{author}{Ming-Wei \surnamestart Chang\surnameend},
  \bibinfo{author}{Kenton \surnamestart Lee\surnameend} \&
  \bibinfo{author}{Kristina \surnamestart Toutanova\surnameend}
  (\bibinfo{year}{2019}): \emph{\bibinfo{title}{{BERT}: Pre-training of Deep
  Bidirectional Transformers for Language Understanding}}.
\newblock In: {\slshape \bibinfo{booktitle}{Proceedings of the 2019 Conference
  of the North {A}merican Chapter of the Association for Computational
  Linguistics: Human Language Technologies, Volume 1 (Long and Short Papers)}},
  \bibinfo{address}{Minneapolis, Minnesota}, pp. \bibinfo{pages}{4171--4186},
  \doi{10.18653/v1/N19-1423}.

\bibitemdeclare{article}{Dzhafarov2015}
\bibitem{Dzhafarov2015}
\bibinfo{author}{Ehtibar~N. \surnamestart Dzhafarov\surnameend},
  \bibinfo{author}{Janne~V. \surnamestart Kujala\surnameend},
  \bibinfo{author}{Víctor~H. \surnamestart Cervantes\surnameend},
  \bibinfo{author}{Ru~\surnamestart Zhang\surnameend} \& \bibinfo{author}{Matt
  \surnamestart Jones\surnameend} (\bibinfo{year}{2016}):
  \emph{\bibinfo{title}{On contextuality in behavioural data}}.
\newblock {\slshape \bibinfo{journal}{Philosophical Transactions of the Royal
  Society A: Mathematical, Physical and Engineering Sciences}}
  \bibinfo{volume}{374}(\bibinfo{number}{2068}), p. \bibinfo{pages}{20150234},
  \doi{10.1098/rsta.2015.0234}.

\bibitemdeclare{article}{EPR}
\bibitem{EPR}
\bibinfo{author}{Albert \surnamestart Einstein\surnameend},
  \bibinfo{author}{Boris \surnamestart Podolsky\surnameend} \&
  \bibinfo{author}{Nathan \surnamestart Rosen\surnameend}
  (\bibinfo{year}{1935}): \emph{\bibinfo{title}{Can Quantum-Mechanical
  Description of Physical Reality Be Considered Complete?}}
\newblock {\slshape \bibinfo{journal}{Phys. Rev.}} \bibinfo{volume}{47}, pp.
  \bibinfo{pages}{777--780}, \doi{10.1103/PhysRev.47.777}.

\bibitemdeclare{article}{firth1957synopsis}
\bibitem{firth1957synopsis}
\bibinfo{author}{John~R \surnamestart Firth\surnameend} (\bibinfo{year}{1957}):
  \emph{\bibinfo{title}{A synopsis of linguistic theory, 1930-1955}}.
\newblock {\slshape \bibinfo{journal}{Studies in linguistic analysis}}.

\bibitemdeclare{article}{harris1954distributional}
\bibitem{harris1954distributional}
\bibinfo{author}{Zellig~S. \surnamestart Harris\surnameend}
  (\bibinfo{year}{1954}): \emph{\bibinfo{title}{Distributional Structure}}.
\newblock {\slshape \bibinfo{journal}{WORD}}
  \bibinfo{volume}{10}(\bibinfo{number}{2-3}), pp. \bibinfo{pages}{146--162},
  \doi{10.1080/00437956.1954.11659520}.

\bibitemdeclare{article}{Howard2014}
\bibitem{Howard2014}
\bibinfo{author}{M.~\surnamestart Howard\surnameend},
  \bibinfo{author}{J.~\surnamestart Wallman\surnameend},
  \bibinfo{author}{V.~\surnamestart Veitch\surnameend} \&
  \bibinfo{author}{J.~\surnamestart Emerson\surnameend} (\bibinfo{year}{2014}):
  \emph{\bibinfo{title}{{Contextuality supplies the 'magic' for quantum
  computation}}}.
\newblock {\slshape \bibinfo{journal}{Nature}}
  \bibinfo{volume}{510}(\bibinfo{number}{7505}), pp. \bibinfo{pages}{351--355},
  \doi{10.1038/nature13460}.
\newblock \eprint{1401.4174}.

\bibitemdeclare{inproceedings}{Levesque2012}
\bibitem{Levesque2012}
\bibinfo{author}{Hector~J. \surnamestart Levesque\surnameend},
  \bibinfo{author}{Ernest \surnamestart Davis\surnameend} \&
  \bibinfo{author}{Leora \surnamestart Morgenstern\surnameend}
  (\bibinfo{year}{2012}): \emph{\bibinfo{title}{The Winograd Schema
  Challenge}}.
\newblock In: {\slshape \bibinfo{booktitle}{Proceedings of the Thirteenth
  International Conference on Principles of Knowledge Representation and
  Reasoning}}, \bibinfo{series}{KR'12}, \bibinfo{publisher}{AAAI Press}, p.
  \bibinfo{pages}{552–561}.

\bibitemdeclare{misc}{Lorenzetal}
\bibitem{Lorenzetal}
\bibinfo{author}{Robin \surnamestart Lorenz\surnameend}, \bibinfo{author}{Anna
  \surnamestart Pearson\surnameend}, \bibinfo{author}{Konstantinos
  \surnamestart Meichanetzidis\surnameend}, \bibinfo{author}{Dimitri
  \surnamestart Kartsaklis\surnameend} \& \bibinfo{author}{Bob \surnamestart
  Coecke\surnameend} (\bibinfo{year}{2021}): \emph{\bibinfo{title}{QNLP in
  Practice: Running Compositional Models of Meaning on a Quantum Computer}},
  \doi{10.48550/arXiv.2102.12846}.

\bibitemdeclare{article}{Mansfield2018}
\bibitem{Mansfield2018}
\bibinfo{author}{S.~\surnamestart Mansfield\surnameend} \&
  \bibinfo{author}{E.~\surnamestart Kashefi\surnameend} (\bibinfo{year}{2018}):
  \emph{\bibinfo{title}{{Quantum Advantage from Sequential-Transformation
  Contextuality}}}.
\newblock {\slshape \bibinfo{journal}{Physical Review Letters}}
  \bibinfo{volume}{121}(\bibinfo{number}{23}), pp. \bibinfo{pages}{1--8},
  \doi{10.1103/PhysRevLett.121.230401}.
\newblock \eprint{1801.08150}.

\bibitemdeclare{misc}{Emeriau2022}
\bibitem{Emeriau2022}
\bibinfo{author}{Damian~Markham \surnamestart
  Pierre-Emmanuel~Emariau\surnameend, Shane~Mansfield} (\bibinfo{year}{2022}):
  \emph{\bibinfo{title}{Corrected Bell and Non-Contextuality Inequalities for
  Realistic Experiments}}.
\newblock \bibinfo{howpublished}{in preparation}.

\bibitemdeclare{article}{Raussendorf2013}
\bibitem{Raussendorf2013}
\bibinfo{author}{R.~\surnamestart Raussendorf\surnameend}
  (\bibinfo{year}{2013}): \emph{\bibinfo{title}{{Contextuality in
  measurement-based quantum computation}}}.
\newblock {\slshape \bibinfo{journal}{Physical Review A - Atomic, Molecular,
  and Optical Physics}} \bibinfo{volume}{88}(\bibinfo{number}{2}), pp.
  \bibinfo{pages}{1--7}, \doi{10.1103/PhysRevA.88.022322}.
\newblock \eprint{0907.5449}.

\bibitemdeclare{inproceedings}{Wang2021a}
\bibitem{Wang2021a}
\bibinfo{author}{Daphne \surnamestart Wang\surnameend},
  \bibinfo{author}{Mehrnoosh \surnamestart Sadrzadeh\surnameend},
  \bibinfo{author}{Samson \surnamestart Abramsky\surnameend} \&
  \bibinfo{author}{Victor \surnamestart Cervantes\surnameend}
  (\bibinfo{year}{2021}): \emph{\bibinfo{title}{On the Quantum-like
  Contextuality of Ambiguous Phrases}}.
\newblock In: {\slshape \bibinfo{booktitle}{Proceedings of the 2021 Workshop on
  Semantic Spaces at the Intersection of NLP, Physics, and Cognitive Science
  (SemSpace)}}, \bibinfo{publisher}{Association for Computational Linguistics},
  \bibinfo{address}{Groningen, The Netherlands}, pp. \bibinfo{pages}{42--52}.

\bibitemdeclare{article}{Wang2021b}
\bibitem{Wang2021b}
\bibinfo{author}{Daphne \surnamestart Wang\surnameend},
  \bibinfo{author}{Mehrnoosh \surnamestart Sadrzadeh\surnameend},
  \bibinfo{author}{Samson \surnamestart Abramsky\surnameend} \&
  \bibinfo{author}{Víctor~H. \surnamestart Cervantes\surnameend}
  (\bibinfo{year}{2021}): \emph{\bibinfo{title}{{Analysing Ambiguous Nouns and
  Verbs with Quantum Contextuality Tools}}}.
\newblock {\slshape \bibinfo{journal}{Journal of Cognitive Science}}
  \bibinfo{volume}{22}(\bibinfo{number}{3}), pp. \bibinfo{pages}{391--420},
  \doi{10.17791/jcs.2021.22.3.391}.

\end{thebibliography}

\section{Appendix}

\subsection{Data for adjectives}
\begin{center}
    \begin{tabular}{|r|p{0.4\textwidth}|c|c|}
        \hline
        noun pair & adjective modifiers & models & contextual models \\ \hline\hline
        cat, dog & cute, furry, lovely, friendly, sweet, big, small, house, young, large, wild, dead, thirsty, hungry, good, gray, black, little & 9792 & 344 \\ \hline
        girl, boy & little, beautiful, young, pretty, small, baby, teenage & 420 & 1 \\\hline
        man, woman & young, dead, little, big, strange, beautiful, tall & 420 & 5 \\\hline
        strawberry, apple & round, red, sweet, sour, rotten & 120 & 0 \\ \hline
        daisy, marigold & yellow, small, beautiful, everywhere & 48 & 0 \\\hline
        daisy, sunflower & yellow, small, beautiful & 12 & 0 \\\hline
        moth, butterfly & winged, colorful, light, beautiful & 48 & 0 \\\hline
        cucumber, courgette & green, long, juicy, tasty & 48 & 0 \\\hline
        dolphin, porpoise & grey, wet, slippery, slim & 48 & 0 \\\hline
        potato, yam & orange, starchy, healthy, big & 48 & 0 \\\hline
        car, bus & fast, sturdy, safe, heavy & 48 & 0 \\ \hline
    \end{tabular}
\end{center}

\subsection{Data for verbs}

 \begin{center}
    \begin{tabular}{|r|p{0.4\textwidth}|c|c|}
        \hline
        noun pair & verbs & models & contextual models \\ \hline\hline
        strawberry, apple & sold, bought, washed, eaten, rotten, cooked, chilled, steamed
& 672 & 9  \\ \hline
        cat, dog & fed, chased, watched, held, hunted, touched, pet, bathed, cleaned & 1008 & 75 \\\hline
    \end{tabular}
\end{center}

\subsection{Data for prepositional phrases}
 \begin{center}
    \begin{tabular}{|r|p{0.4\textwidth}|c|c|}
        \hline
        noun pair & prepositional phrase modifiers & models & contextual models \\ \hline\hline
        apple, strawberry & on the table, in a dish, in the fridge & 12 & 1 \\ \hline
        boy, girl & from the town, at the school, near the shop, on a bus, across the street, in the city & 240 & 1  \\ \hline
    \end{tabular}
\end{center}

\end{document}